\documentclass[a4paper, conference]{IEEEtran}
\usepackage[style=ieee, url=false, doi=false, isbn=false]{biblatex}
\usepackage{graphicx}
\usepackage{mathtools}
\usepackage{siunitx}
\usepackage{multirow}
\usepackage{xspace}
\usepackage{array}
\usepackage{makecell}
\usepackage{xparse}
\usepackage[bookmarks=false]{hyperref}
\usepackage{amsmath}
\usepackage{amssymb}
\usepackage{fontawesome5}
\usepackage{booktabs}
\usepackage{comment}
\usepackage{placeins}
\usepackage{tikz}
\usetikzlibrary{
  positioning,
  fit,
  arrows.meta,
  backgrounds,
  shapes.geometric,
  calc
}
\newcommand{\name}{LatentCSI\xspace}
\newcommand{\sdimgscale}{0.25\linewidth}

\renewcommand{\bfseries}{\fontseries{b}\selectfont}
\robustify\bfseries
\newrobustcmd{\B}{\bfseries}

\newcommand{\samplefigs}[2]{%
    \includegraphics[width=0.2\linewidth]{./figures/#1/samples/r_#2.png} &
    \includegraphics[width=0.2\linewidth]{./figures/#1/samples/p_#2.png} &
    \includegraphics[width=0.2\linewidth]{./figures/#1/samples/b_#2.png} &
    \includegraphics[width=0.2\linewidth]{./figures/#1/samples/g_#2.png} \\
}

\IEEEoverridecommandlockouts
\title{High-resolution efficient image generation from WiFi CSI using a pretrained
  latent diffusion model

  \thanks{This work was supported by JSPS KAKENHI Grant Number 23K24831 and
23K26109, and JST ASPIRE Grant Number JPMJAP2346, Japan. Models were
trained on the TSUBAME4 supercomputer at the Institute of Science
Tokyo.}
}

\author{ \IEEEauthorblockN{Eshan Ramesh, Takayuki Nishio}
  \IEEEauthorblockA{School of Engineering, Institute of Science Tokyo \\
    Email: esrh@esrh.me, nishio@ict.eng.isct.ac.jp
  }}

\begin{document}
\maketitle

\begin{abstract}
We present LatentCSI, a novel method for generating images of the physical
environment from WiFi CSI measurements that leverages a pretrained latent
diffusion model (LDM).  Unlike prior approaches that rely on complex and
computationally intensive techniques such as GANs, our method employs a
lightweight neural network to map CSI amplitudes directly into the latent
space of an LDM. We then apply the LDM’s denoising diffusion model to the
latent representation with text-based guidance before decoding using the LDM’s
pretrained decoder to obtain a high-resolution image. This design bypasses the
challenges of pixel-space image generation and avoids the explicit image
encoding stage typically required in conventional image-to-image pipelines,
enabling efficient and high-quality image synthesis. We validate our approach
on two datasets: a wide-band CSI dataset we collected with off-the-shelf WiFi
devices and cameras; and a subset of the publicly available MM-Fi dataset. The
results demonstrate that LatentCSI outperforms baselines of comparable
complexity trained directly on ground-truth images in both computational
efficiency and perceptual quality, while additionally providing
practical advantages through its unique capacity for text-guided controllability.
\end{abstract}

\section{Introduction}

WiFi channel state information is typically computed in the process of
equalizing received symbols using known sync words and pilot carriers. However,
especially in multi-carrier systems, it contains rich information about the
propagation environment. The ubiquity of WiFi therefore makes CSI an attractive
source of data for sensing applications. Data-driven methods for recovering
physical information from the CSI of a multi-carrier wireless link are well
studied. With the power of machine learning techniques, researchers have
succeeded in accomplishing a wide variety of sensing tasks, including pose
estimation and activity classification \cite{ma_wifi_2019}, and with sufficient
bandwidth, even as finely-grained tasks as sensing human breathing at a
distance. While existing research has confirmed the feasibility, and indeed the
surprising practicality, of using CSI to predict low-dimensional indicators of
the physical world, work showing high-dimensional environment reconstruction
capability is comparatively scarce.

In this work, we are interested in generating high resolution RGB camera images
from CSI. In the human sensing case (which is a primary focus of the CSI sensing
domain) camera images effectively superset the information gained from pose
estimation and activity classification. Image generation is considerably more
complex, demanding good perceptual structure and high output dimensionality.

Given these challenges, many prior works have adopted generative adversarial
networks (GANs) \cite{goodfellow_generative_2014} as a solution for CSI-based
image generation.  The GAN framework trains two models against each other: a
generator that learns the target distribution and a discriminator that learns to
distinguish generated samples from real samples. GAN-based approaches were some
of the first to be applied to generating images from CSI; one of the first
direct CSI to image works \cite{kato_csi2image_2021} developed a GAN to
generate a 64x64 RGB image. The authors demonstrated their method correctly
classifying objects and positioning human subjects, but they noted poor
performance on humans in motion. Subsequent work in
\cite{yu_wifi-based_2022} and \cite{yu_rfgan_2023} applies the GAN framework to
transform images using information determined from CSI
using pose keypoints determined from CSI, which circumvents the challenge of
generating images from scratch. While GANs have shown promise, they introduce significant complexity into
the training process, as they require training two models simultaneously and
careful tuning to ensure stability and convergence
\cite{arjovsky_towards_2017}. This complexity motivates the exploration of
simpler models with more straightforward training procedures. A recent non-GAN
approach addressing this issue is presented in
\cite{strohmayer_through-wall_2024}, where the authors propose a
mixture-of-product-of-experts variational autoencoder (VAE) that jointly trains
an MLP-based CSI encoder and a ground-truth image encoder.

Despite these efforts, the existing literature remains limited in image
resolution and computational efficiency. Moreover, to our knowledge, the
existing methods—including both GAN-based and non-GAN approaches—rely on
end-to-end training with ground-truth images. As a result, the model tends to
fit the image data very well, leading to generated outputs that inadvertently
reproduce fine visual features. This is particularly concerning in
privacy-sensitive scenarios, where unintended reconstruction of details such as
facial features may pose significant privacy risks.

To address these limitations, we propose \name, a novel image generation scheme
that leverages a pretrained latent diffusion model (LDM). Diffusion models have
recently gained attention for their ability to produce high-quality images
through iterative denoising guided by text embeddings. Latent diffusion models
in particular operate in a lower-dimensional latent space rather than the color
pixel space, enabling faster and more memory-efficient image synthesis.

Compared to existing methods, \name adopts a simpler design that trains a
single, lightweight neural network to predict latent embeddings directly from
CSI data. The reduced dimensionality of the output, combined with the expressive
power of a pretrained LDM diffusion model and decoder allows for the generation of
high-resolution images at a lower computational cost. Moreover, since the LDM
supports text-guided image generation, our approach naturally enables an
optional refinement step, in which the generated image can be modified according
to textual instructions—for example, altering a person’s face or clothing
appearance to match a given description. Importantly, the low-dimensional
bottleneck in our encoding-decoding pipeline inherently reduces the risk of
reproducing sensitive fine-grained details such as facial features, clothing, or
background elements. At the same time, the LDM can synthesize visually plausible
content aligned with user intent via text-guided modification, striking a
balance between interpretability and privacy. This capability enables
controllable and privacy-preserving image generation while preserving a
meaningful correspondence with the underlying CSI signals.

The primary contribution of this work is the introduction of an LDM-based
framework for generating images of the physical environment from CSI data. Our
approach achieves state-of-the-art resolution and text-conditioned image
synthesis with a simple architecture, a straightforward training procedure, and
low computational overhead. We validate our method using two paired CSI–image
datasets: one collected through experiments involving variations in a walking
subject's position and orientation; and another based on a subset of the
publicly available MM-Fi dataset \cite{yang_mm-fi_2023}, selected to represent
diverse human poses with less visual complexity. For both datasets, we
qualitatively evaluate the generated images and quantitatively compare our
approach against a baseline model trained end-to-end on
full-resolution images and the hybrid GAN strategy introduced in
CSI2Image \cite{kato_csi2image_2021}.

\section{Methodology}
\subsection{System model}

We consider a WiFi sensing system designed to monitor indoor environments by
capturing human activity through wireless signals. The system consists of a WiFi
access point, a WiFi terminal, and an RGB camera that is available only during
the training phase. This setup enables the collection of paired channel state
information (CSI) and image data, which are used to train an image generation
model. \name, is intended for practical scenarios in which
camera imagery can be collected temporarily—such as during initial system
calibration—but becomes unavailable during deployment due to obstructions or
privacy concerns. As a result, while the model leverages visual data during
training, inference is performed exclusively using CSI, making the approach
suitable for privacy-sensitive or camera-limited environments.

For the approach to be feasible, it is essential that the CSI measurements carry
sufficient information to represent the perceptual variations present in the
visual scene. This requires a strong correspondence between
fluctuations in CSI and changes in image content—i.e., sufficient mutual
information between the two modalities. This assumption, while often
implicit, is fundamental to all prior work on CSI-based image generation. If
the perceptual alignment or the expressive capacity of CSI is lacking,
learning a reliable mapping from CSI to visual content becomes fundamentally
difficult.

Finally, we assume access to a publicly available pretrained latent
diffusion model, such as Stable Diffusion, which provides a complete pipeline
composed of an encoder for mapping RGB images into a latent space, a decoder for
reconstructing images from latent representations, and a denoising diffusion
model for generating and refining latent representations. The availability and
perceptual quality of this fully pretrained LDM is essential to our approach,
as it enables high-quality image synthesis in a computationally efficient
manner, eliminating the need to train large generative models from scratch.

\subsection{Overview of \name}

Our proposed method, \name, builds upon a pretrained latent diffusion model,
Stable Diffusion v1.5 \cite{rombach_high-resolution_2022}.  To clarify our
design, we first describe the standard inference process of LDMs. We proceed to
introduce our modified inference procedure, which adapts an image-to-image
pipeline to use CSI instead of images. The training procedure of the key
component in this pipeline---the CSI encoder that predicts the LDM's latent
representation---is presented in the following section.

\subsubsection{LDM image-to-image pipeline}

Diffusion models generate images by reversing a diffusion process, progressively
transforming sampled Gaussian noise into an image through iterative denoising
guided by conditioning inputs such as text prompts. In contrast, the
image-to-image procedure, popularized in \cite{meng_sdedit_2021}, starts from an
input image, adds noise to simulate forward diffusion up to a chosen
intermediate timestep, and then denoises to produce the output.

\begin{figure}[htbp]
  \centering
  \begin{tikzpicture}[node distance=0.8cm and 0.3cm, framed, font=\footnotesize]
    \tikzset{ dnn/.style={draw, trapezium, shape border rotate=270, fill=green!30}}

    \node[draw, rectangle, align=center] (wifi) {\faVideo\ Images\\($y$)};
    \node[draw, trapezium, shape border rotate=270, fill=red!40, right=of wifi]
    (dnn) {\rotatebox{90}{VAE Encoder}};

    \node[draw, rectangle, right=of dnn, fill=red!10] (p) {$z$};

    \node[draw, rectangle, rounded corners, right=of p, minimum
    height=1.5cm, fill=red!40] (ldm) {LDM};
    \node[draw, rectangle, above=of ldm] (text) {\faCommentDots\ Text
      conditioning};

    \coordinate (mid) at ($(p.east)!0.5!(ldm.west)$);
    \node[draw, rectangle, fill=gray!10] at ($(mid) + (0,-1.5)$) (noise)
    {$n = \mathcal{N}(0, \sigma^2)$};
    \draw [-] (noise) -- (mid);

    \node[draw, rectangle, right=of ldm] (out_lat)
    {$\text{LDM}(z + n)$};

    \node[draw, trapezium, shape border rotate=90, fill=red!40,
    right=of out_lat] (dec) {\rotatebox{90}{VAE Decoder}};

    \node[draw, right=of dec] (output) {Output};

    \draw[->] (wifi) -- (dnn); \draw[->] (dnn) -- (p); \draw[->] (p) -- (ldm);
    \draw[->] (text) -- (ldm); \draw[->] (ldm) -- (out_lat);
    \draw[->] (out_lat) -- (dec); \draw[->] (dec) -- (output);

  \end{tikzpicture}
  \caption{Stable diffusion LDM img2img pipeline}
  \label{fig:sysmod_typical}
\end{figure}
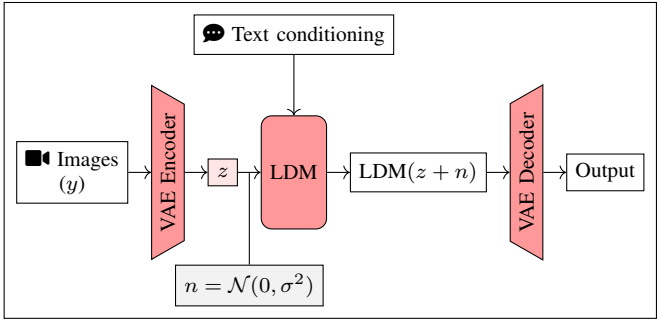

Latent diffusion models perform denoising in latent space rather than pixel
space, offering performance benefits. Since our implementation of \name is based
on Stable Diffusion, we describe the LDM's operation according to the
implementation of Stable Diffusion. While different implementations of LDM may
vary in model architecture, their fundamental operation remains the same. The
image-to-image pipeline of an LDM is illustrated in
\autoref{fig:sysmod_typical}.

A VAE with Kullback–Leibler (KL) regularization, trained on large
datasets to preserve perceptual structure under compression, serves as the
encoder and decoder to map between image and latent spaces. The denoising
diffusion model, a U-Net, is trained to estimate and remove noise at each
diffusion timestep, conditioned on text embeddings.

In the LDM image-to-image process, the input image is first encoded into a
latent embedding, to which a parametrized amount of noise is added. This noisy
latent embedding is then iteratively denoised by the denoising diffusion model,
guided by conditioning inputs (e.g., a text prompt). Finally, the decoder maps
the denoised latent embedding back to image space to generate the target image.

The variance of noise added to the latent vectors before they are passed through
the denoising diffusion model, and therefore the intermediate timestep the
reverse process begins at, is an important parameter, typically called
``strength''. The closed-form expression for the noise variance depends on the
diffusion algorithm and variance schedule used. Lower strength corresponds to
output images that more closely resemble the input, and higher strength
induces more significant transformations.

\subsubsection{\name Inference Process}

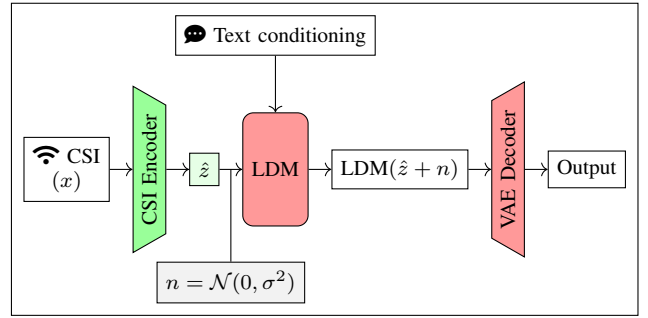
\begin{figure}[htbp]
  \centering
  \begin{tikzpicture}[node distance=0.75cm and 0.3cm, framed, font=\footnotesize]
    \tikzset{ dnn/.style={draw, trapezium, shape border rotate=270,
        fill=green!30} }

    \node[draw, rectangle, align=center] (wifi) {\faWifi\ CSI\\$\left(x\right)$};

    \node[draw, trapezium, shape border rotate=270, fill=green!40,
    right=of wifi] (dnn) {\rotatebox{90}{CSI Encoder}};
    \node[draw,
    rectangle, right=of dnn, fill=green!10] (p) {$\hat{z}$};

    \node[draw, rectangle, rounded corners, right=of p, minimum
    height=1.5cm, fill=red!40] (ldm) {LDM};
    \node[draw, rectangle,
    above=of ldm] (text) {\faCommentDots\ Text conditioning};

    \node[draw, rectangle, right=of ldm] (out_lat)
    {$\text{LDM}(\hat{z} + n)$};
    \node[draw, trapezium, shape border
    rotate=90, fill=red!40, right=of out_lat] (dec) {\rotatebox{90}{VAE Decoder}};
    \node[draw, right=of dec] (output) {Output};

    \coordinate (mid) at ($(p.east)!0.5!(ldm.west)$);
    \node[draw, rectangle, fill=gray!10] at ($(mid) + (0,-1.5)$) (noise) {$n = \mathcal{N}(0, \sigma^2)$};
    \draw [-] (noise) -- (mid);

    \draw[->] (wifi) -- (dnn); \draw[->] (dnn) -- (p); \draw[->] (p) --
    (ldm); \draw[->] (text) -- (ldm); \draw[->] (ldm) -- (out_lat);
    \draw[->] (out_lat) -- (dec); \draw[->] (dec) -- (output);

  \end{tikzpicture}
  \caption{\name: img2img with custom CSI encoder}
  \label{fig:sysmod_inference}
\end{figure}

\autoref{fig:sysmod_inference} shows the inference pipeline of \name. The key
difference from the standard LDM pipeline is that the VAE encoder is replaced
with a CSI encoder, enabling the LDM’s latent starting point to be estimated
directly from CSI instead of an image. Similar to an LDM, data in the \name
pipeline is processed primarily in latent space, in contrast to existing
CSI-to-image generation methods that directly map CSI to image space. This
latent-space design reduces computational cost, as the latent embedding has much
lower dimensionality than high-resolution images.

The CSI encoder plays a crucial role by mapping CSI
measurements into the latent space expected by the LDM's denoising model. The
accuracy of this latent embedding is vital, as it serves as the input for the
subsequent denoising and decoding stages. Once the latent embedding is estimated
by the CSI encoder, it is iteratively denoised by the pretrained denoising
diffusion model under text-based conditioning, and finally decoded into an image
using the pretrained decoder.

We emphasize that the only modified component in this otherwise standard LDM
pipeline is the custom CSI encoder. As a result, training is straightforward and
computationally light, since only the CSI encoder needs to be fitted while all
other components can be used without finetuning. The training procedure for the
CSI encoder is detailed in the following section.

\subsection{\name training process}

The CSI encoder is trained to map CSI measurements into the latent space of the
LDM as a probabilistic model. Specifically, let \( x \) denote the CSI input and
\( y \) the corresponding image. We model the behavior of the CSI encoder as a
probabilistic distribution \( p_w(z|x) \), parameterized by \( w \), analogous
to the pretrained VAE encoder \( q_{\Phi}(z|y) \) of the LDM. The goal is to train
\( p_w(z|x) \) so that its output distribution closely matches
\( q_{\Phi}(z|y) \). This is achieved by minimizing the divergence between the
latent samples from \( p_w(z|x) \) and \( q_{\Phi}(z|y) \) over paired training
data. In other words, the CSI encoder learns to approximate the latent
distribution produced by the pretrained VAE encoder for the corresponding
image.

The pretrained VAE encoder \( q_{\Phi}(z|y) \) models the posterior latent
distribution as a Gaussian with a diagonal covariance matrix:
\[ q_{\Phi}(z|y) = \mathcal{N}(\mu_{\Phi}(y), \mathrm{diag}(\sigma_{\Phi}^2(y))) \] with
\(\mu_{\Phi}(y), \sigma_{\Phi}^2(y) \in \mathbb{R}^{4 \times 64 \times 64} \).  To obtain a sample
\( z \), we draw from the distribution using the reparameterization trick:
$z = \mu + \sigma \odot \epsilon, \quad \epsilon \sim \mathcal{N}(0, I)$.  In the VAE of Stable Diffusion v1.5, the
predicted variance \( \sigma_\Phi^2(y) \) has been empirically observed to be very
small, resulting in sampled latent vectors \( z \) concentrating closely around
the mean. Therefore, we directly use \( \mu_\Phi(y) \) as the latent embedding:
$z \coloneq \mu_\Phi(y).$

As a result of this choice, we redefine the CSI encoder as a deterministic neural network $f_w(x)$, which predicts the latent mean $\mu_\Phi(y)$ from a CSI measurement $x$, replacing the earlier probabilistic formulation.
The detailed architecture of the CSI encoder is described in
the following subsection. The CSI encoder is trained by minimizing the squared
error between its output and the corresponding latent mean:
\[ \min_{w} \; \mathbb{E}_{(x,y) \sim \mathcal{D}} \left[ \| f_w(x) - \mu_\Phi(y) \|_2^2 \right].
\]
Here, \(\mathcal{D}\) denotes the empirical joint distribution of paired training samples
\((x, y)\), where \(x\) is a CSI measurement and \(y\) is the corresponding RGB
image captured simultaneously during data collection.

\begin{figure}[htbp]
  \centering
  \begin{tikzpicture}[node distance=2cm and 0.5cm, framed, font=\footnotesize]
    \node[draw, rectangle] (cam) {\faVideo\ Images ($y$)};
    \node[draw, trapezium, shape border rotate=270, fill=red!40,
    right=1cm of cam]
    (vae) {\rotatebox{90}{VAE Encoder}};

    \node[draw, rectangle, right=0.5cm of vae, fill=red!10] (y) {Latent samples ($z$)};

    \node[draw, rectangle, below=1.5cm of cam.west, anchor=west] (wifi)
    {\faWifi\ CSI ($x$)};
    \node [draw, trapezium, shape border rotate=270, fill=green!40, right=of wifi] (dnn)
    {\rotatebox{90}{CSI Encoder}};
    \node[draw, rectangle, right=of dnn, fill=green!10] (p) {$\hat{z}$};

    \node[draw, ellipse] (loss) at ([yshift=-1.5cm]y) {$L(\hat{z}, z)$};

    \draw[->] (cam) -- (vae);
    \draw[->] (vae) -- (y);
    \draw[->] (wifi) -- (dnn);
    \draw[->] (dnn) -- (p);
    \draw[->] (p) -- (loss);
    \draw[->] (y) -- (loss);
    \coordinate (lowerbound) at ($(loss.south)+(0,-0.75)$);
    \draw[->] (loss.south) |- (lowerbound) -| (dnn.south);

  \end{tikzpicture}
  \caption{CSI encoder training procedure}
  \label{fig:sysmod_training}
\end{figure}
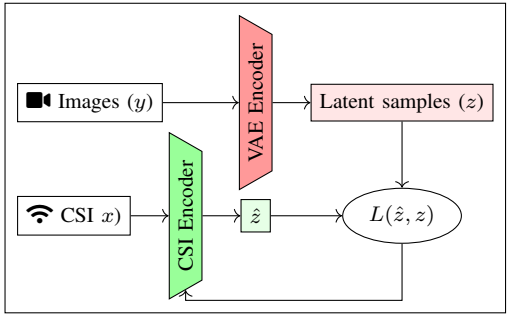

The overall training procedure is illustrated in
\autoref{fig:sysmod_training}. The CSI encoder learns to regress directly to the
mean latent representation produced by the pretrained VAE encoder for each
corresponding image \( y \).  In our approach, the CSI input \( x \) uses only
amplitude information. Given a complex-valued CSI measurement
\( x_c \in \mathbb{C}^s \), the input is computed as
\( x = \sqrt{\text{Re}(x_c)^2 + \text{Im}(x_c)^2} \). This choice follows prior
work \cite{ma_wifi_2019} \cite{yu_wifi-based_2022}
\cite{strohmayer_through-wall_2024} suggesting that amplitude alone is
sufficient for many sensing tasks, and avoids the additional complexity of
handling phases in unsynchronized device configurations.

\subsection{Model architecture}

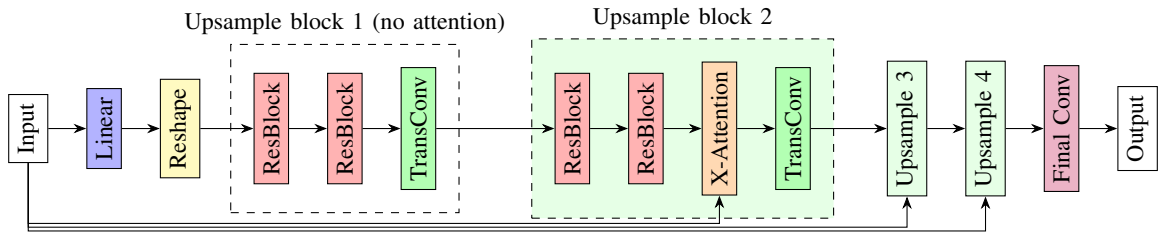
\begin{figure*}
  \centering
  \begin{tikzpicture}[>=Stealth, every node/.style={font=\small}, node
    distance=0.5cm]

    \tikzset{
      input/.style={draw, rectangle, align=center},
      linear/.style={draw, rectangle, fill=blue!30, align=center},
      reshape/.style={
        draw, rectangle, fill=yellow!30, align=center
      },
      resblock/.style={draw, rectangle,
        fill=red!30, align=center
      },
      attention/.style={
        draw, rectangle, fill=orange!30, align=center
      },
      transpose/.style={
        draw, rectangle, fill=green!30, align=center
      },
      finalconv/.style={
        draw, rectangle, fill=purple!30, align=center},
      upsblocknoatt/.style={draw, dashed, rectangle, inner sep=0.3cm},
      upsblock/.style={draw, dashed, rectangle, inner sep=0.3cm, fill=green!10},
      upsample/.style={draw, rectangle, fill=green!10, align=center}
    }

    \node (Input) [input] {\rotatebox{90}{Input}};
    \node (Linear) [linear, right=of Input] {\rotatebox{90}{Linear}};
    \node (Reshape) [reshape, right=of Linear] {\rotatebox{90}{Reshape}};

    \node (res1a) [resblock, right=1.5cm of Reshape, xshift=-0.8cm] {\rotatebox{90}{ResBlock}};
    \node (res1b) [resblock, right=of res1a] {\rotatebox{90}{ResBlock}};
    \node (trans1) [transpose, right=of res1b] {\rotatebox{90}{TransConv}};
    \node (Block1) [upsblocknoatt, fit=(res1a) (res1b) (trans1), label=above:{Upsample block 1 (no attention)}] {};

    \node (res2a) [resblock, right=2.25cm of Block1, xshift=-1.0cm] {\rotatebox{90}{ResBlock}};
    \node (res2b) [resblock, right=of res2a] {\rotatebox{90}{ResBlock}};
    \node (att2) [attention, right=of res2b] {\rotatebox{90}{X-Attention}};
    \node (trans2) [transpose, right=of att2] {\rotatebox{90}{TransConv}};

\begin{scope}[on background layer]
  \node (Block2) [upsblock, fit=(res2a) (res2b) (att2) (trans2),
  label=above:{Upsample block 2}] {};
\end{scope}

\node (ups3) [upsample, right=1cm of trans2] {\rotatebox{90}{Upsample 3}};
\node (ups4) [upsample, right= of ups3] {\rotatebox{90}{Upsample 4}};

\node (Final) [finalconv, right=of ups4] {\rotatebox{90}{Final Conv}};

\node (out) [draw, rectangle, right= of Final] {\rotatebox{90}{Output}};

\draw[->] (Input) -- (Linear); \draw[->] (Linear) -- (Reshape);
\draw[->] (Reshape) -- (res1a);
\draw[->] (trans1) -- (res2a);
\draw[->] (trans2) -- (ups3);
\draw[->] (ups3) -- (ups4);
\draw[->] (ups4) -- (Final);
\draw[->] (Final) -- (out);

\draw[->] (res1a) -- (res1b);
\draw[->] (res1b) -- (trans1);
\draw[->] (res2a) -- (res2b);
\draw[->] (res2b) -- (att2);
\draw[->] (att2) -- (trans2);

\draw[->] (Input.south) -- ($(Input.south)+(0,-0.8)$) -| (att2.south);
\draw[->] (Input.south) -- ($(Input.south)+(0,-0.85)$) -| (ups3.south);
\draw[->] (Input.south) -- ($(Input.south)+(0,-0.9)$) -| (ups4.south);
\end{tikzpicture}
\caption{Diagram of CSI encoder model}
\label{fig:enc_arch}
\end{figure*}

Our model architecture is shown in \autoref{fig:enc_arch}. The input to the
model is the preprocessed amplitude of the CSI signal, and the
output is a latent embedding in the latent space of the pretrained
VAE in the LDM pipeline.

We first apply a fully connected layer to map the 1D CSI input of size
$s$ to a 1D vector before reshaping it into a tensor with $b$
channels, where $b$ is a tunable parameter. We use 4 upsampling steps
that each halves the number of channels and doubles the resolution. We
then use a final convolutional layer to reduce the number of channels
to the output specification. Each upsampling layer consists of two
residual blocks and a transpose convolutional layer. Each residual
block is itself composed of two convolutional layers. The last 3
upsampling steps use a cross attention layer to ensure that later
spatially upsampling blocks attend to global information. This
architecture strongly resembles that of the Stable Diffusion VAE
encoder, which uses 4 similar downsampling steps followed by
convolutional layers.

\section{Evaluation}

We evaluate \name on two datasets. For each dataset, we present a qualitative
analysis using sample reference and generated images from the test sets withheld
from the model.

To quantitatively assess the performance of the proposed method, we compare
ground-truth images with predicted images without the application of latent
diffusion transformations—specifically, the noising strength parameter in the
inference pipeline is set to zero. This ensures that the output
remains faithful to the direct predictions of the CSI encoder for an
interpretable and unbiased evaluation.

We use three metrics in our analysis: the pixel-wise Root Mean
Square Error (RMSE), the Structural Similarity Index Measure (SSIM), and the
Fr\'echet Inception Distance (FID). RMSE and SSIM provide image-level
comparisons and are sensitive to absolute and structural differences,
respectively. In contrast, FID offers a perceptual measure of distributional
similarity.

For both datasets, we report these metrics in two cases: (1) using the entire
generated and reference images, and (2) using cropped image regions containing
human figures, as identified by the YOLOv3 object detection model
\cite{redmon_yolov3_2018}. Given the human-centric nature of the datasets, the
second case provides a focused evaluation of the model's ability to
preserve semantically important content.

In both evaluation schemes, we benchmark \name against two strategies:
a baseline model using the exact same architecture and training
procedure as LatentCSI but predicting pixels directly; and the
``hybrid'' GAN strategy (K=8) from CSI2Image\cite{kato_csi2image_2021}
which combines supervised loss with adversarial conditional GAN loss.

\name and the baseline model were trained using an early stopping criterion
for validation loss stagnation over 5 consecutive training epochs. We used the
ADAM optimizer with a fixed learning rate of 0.0005. To account for variability
arising from stochastic weight initialization and training dynamics, each model
was trained 5 times. Image samples are generated using the model instance with
the best test loss. All experiments were conducted on a system equipped with a
single NVIDIA H100 GPU and eight CPU cores of an AMD EPYC 9654 processor.

\subsection{Dataset 1: Position \& orientation of one moving human subject}

We first evaluate the proposed method on a custom dataset designed to assess
performance in predicting the position and orientation of a moving human
subject. Data was collected from a single individual walking back and forth in a
small indoor environment. CSI measurements were recorded at 10~Hz over 25
minutes using two ASUS NUCs equipped with Intel AX210 wireless cards, operating
over a 160~MHz channel with a single transmit-receive antenna pair. A total of
1992 subcarriers were captured per sample using the FeitCSI tool
\cite{feitcsi:project}. Each CSI sample was paired with a synchronized
$640 \times 480$ RGB image captured via an Intel Realsense D435 camera. The resulting
15{,}000 image-CSI pairs were post-processed by cropping and resampling images
to $512 \times 512$ resolution. Facial features were blurred only in the presented
examples; unaltered images were used for training and inference. The dataset was
split into 80\% training, 10\% validation, and 10\% testing. The CSI encoder
model was configured with $b = 256$ initial channels, yielding $22{,}914{,}052$
trainable parameters. The baseline model used $b = 8$ and comprised
$16{,}434{,}395$ parameters.

\begin{table}
  \centering
  \caption{Sample image comparison from dataset 1}
  \label{tab:d1_samples}
  \setlength{\tabcolsep}{1pt}
  \begin{tabular}{cccc}
    \toprule
    Reference & LatentCSI & Baseline & CSI2Image\cite{kato_csi2image_2021} \\
    \midrule
    \samplefigs{walking}{0}
    \samplefigs{walking}{1}
    \samplefigs{walking}{2}
    \samplefigs{walking}{3}
    \bottomrule
  \end{tabular}
\end{table}

\autoref{tab:d1_samples} presents representative samples from the test set,
while quantitative results are shown in \autoref{tab:quant_metrics}. \name
exhibits marginal improvements over the baseline in RMSE and SSIM, but achieves
a substantially lower FID, indicating superior perceptual quality. This
advantage is further amplified when evaluation is restricted to regions
containing human figures, suggesting better reconstruction of salient
information. Despite comparable model complexity, the baseline model—which
directly predicts pixel values—requires approximately three times longer to
train, owing to the increased output dimensionality. Dataset 1 presents a
realistic, unsanitized setting with diverse views of a moving subject and
frequent occlusions. These factors
increase pose significant challenges for pixel-space prediction.

Qualitative results indicate that the baseline and CSI2Image underfit
the dataset, likely constrained by their limited capacity to model
high-resolution pixel distributions. In contrast, \name produces
better reconstructions with a smaller model. These outputs serve as
effective initialization points for downstream diffusion-based
refinement, as shown in \autoref{tab:sd}.

\subsection{Dataset 2: Human poses with MM-Fi}
We evaluate our proposed method on the popular MM-Fi dataset
\cite{yang_mm-fi_2023}. Data is collected from 3 antennas each using 114-carrier
40~MHz channels, resulting in 342 total subcarriers. Compared to Dataset~1, this
offers lower spectral resolution but greater spatial diversity. The selected
subset comprises environment~3, the first two subjects, and four arm movement
activities: 13–14 (raising hand, left/right) and 17–18 (waving hand,
left/right), totaling 23{,}760 samples. The proposed model is configured with
$b = 256$ and has 13{,}621{,}252 parameters; the baseline uses $b = 32$ with
11{,}490{,}275 parameters.

\begin{table}
  \centering
  \caption{Sample image comparison from dataset 2}
  \label{fig:d2_samples}
  \setlength{\tabcolsep}{1pt}
  \begin{tabular}{cccc}
    \toprule
    Reference & LatentCSI & Baseline & CSI2Image\cite{kato_csi2image_2021}\\
    \midrule
    \samplefigs{mmfi}{0}
    \samplefigs{mmfi}{1}
    \samplefigs{mmfi}{2}
    \samplefigs{mmfi}{3}
    \bottomrule
  \end{tabular}
\end{table}

\autoref{fig:d2_samples} and \autoref{tab:quant_metrics} present qualitative and
quantitative results, respectively. Unlike Dataset~1, the visual consistency of
MM-Fi—fixed subject position and minimal background variation—enables the
baseline model to perform well, particularly in terms of RMSE and SSIM. This
reflects its ability to memorize low-variance visual features through direct
pixel supervision.

However, the proposed method outperforms the baseline in FID,
indicating superior perceptual quality. When evaluation is restricted
to regions containing human subjects, the RMSE and SSIM differences
narrow, while the FID advantage of the proposed model approximately
doubles. Qualitative analysis suggests the baseline model tends to
produce artifacts and anatomically implausible outputs. The CSI2Image
model is significantly underfit, and likely requires architecture
tuning to work well with the higher resolution used in this work. In
contrast, the latent-space model yields more coherent representations
of salient features.

\begin{table*}[htbp]
  \centering
  \caption{Quantitative comparison of methods over 5 runs}
\sisetup{
  separate-uncertainty = true,
  detect-weight = true,
  detect-mode = true,
  table-number-alignment = center,
  round-mode=places,
  round-precision=2,
  retain-zero-uncertainty=true
}
\resizebox*{\textwidth}{!}{
\resizebox*{\textwidth}{!}{
\begin{tabular}{
  l
  S[table-format=3.2(2)]  
  S[table-format=2.2(2)]  
  S[table-format=1.2(2)]  
  S[table-format=3.2(2)]  
  S[table-format=2.2(2)]  
  S[table-format=1.2(2)]  
  S[table-format=2.1(1)]  
  l                      
}
\toprule
Method
& {FID $\downarrow$}
& {RMSE $\downarrow$}
& {SSIM $\uparrow$}
& {FID (crop) $\downarrow$}
& {RMSE (crop) $\downarrow$}
& {SSIM (crop) $\uparrow$}
& {Sec/epoch}
& Time (min:s) \\
\midrule

Dataset 1 \name
& \B 134.23(7.18)
& \B 18.95(0.44)
& \B 0.87(0.00)
& \B 126.09(7.15)
& \B 19.81(0.66)
& \B 0.83(0.01)
& \B 11.9(0.1)
& \B $\text{05:02} \pm \text{00:46}$ \\

Dataset 1 Baseline
& 268.03(4.23)
& 20.45(0.27)
& 0.84(0.00)
& 296.93(4.83)
& 21.74(0.29)
& 0.78(0.00)
& 36.4(7.9)
& $\text{16:29} \pm \text{03:35}$ \\

Dataset 1 CSI2Image
& 392.46(11.0)
& 39.14(1.94)
& 0.58(0.06)
& 349.87(15.87)
& 45.57(2.47)
& 0.49(0.04)
& 67.8(1.7)
& $\text{15:28} \pm \text{5:09}$ \\

\midrule

Dataset 2 \name
& \B 28.21(0.78)
& 7.87(0.09)
& 0.94(0.00)
& \B 27.90(0.43)
& 7.90(0.03)
& 0.92(0.00)
& \B 17.9(0.2)
& \B $\text{16:24} \pm \text{00:03}$ \\

Dataset 2 Baseline
& 47.67(2.04)
& \B 7.15(0.50)
& \B 0.97(0.00)
& 69.34(2.50)
& \B 7.19(0.24)
& \B 0.93(0.00)
& 59.3(0.2)
& $\text{91:09} \pm \text{19:19}$ \\

Dataset 2 CSI2Image
& 312.24(21.5)
& 24.15(0.64)
& 0.76(0.03)
& 372.99(10.25)
& 49.72(0.65)
& 0.46(0.02)
& 73.0(2.3)
& $\text{32:53} \pm \text{10:09}$ \\

\bottomrule
\end{tabular}
}
}
 \label{tab:quant_metrics}
\end{table*}

\subsection{Text-guided CSI Image Generation}

Our choice of latent space as that of the Stable Diffusion LDM enables
text-guided manipulation of generated images without requiring an image
encoding step. This capability allows us to refine or alter the
visual content of CSI-generated latent representations through natural language
prompts. As illustrated in \autoref{tab:sd}, \name can generate stylistic
and semantic variations of an image from the same CSI input using
different text prompts. For the results shown, we employ 100 denoising steps of DDIM \cite{song_denoising_2020} with a noise strength of 0.6 which determines the variance of noise added.

Due to the dimensionality bottleneck in \name, high-frequency
details—such as facial features, fine textures, or text—are not preserved.
Text-based conditioning allows us to reconstruct or augment these elements
according to a specified prompt, enabling stylistic transformation
and controlled editing of semantically relevant content.

\begin{table}
  \centering
  \caption{Text-guided CSI Image Generation}
  \label{tab:sd}
  \begin{tabular}{cccc}
    \toprule
    & Reference
    & {\tiny\makecell{``a photograph of a man in a\\ small office room, 4k, realistic''}}
    & {\tiny\makecell{``a drawing of a man in a \\ laboratory, anime, 4k''}} \\

    \midrule

    \raisebox{0.5\dimexpr\sdimgscale}[0pt][0pt]{\rotatebox[origin=c]{90}{Dataset 1}} &
    \includegraphics[width=\sdimgscale]{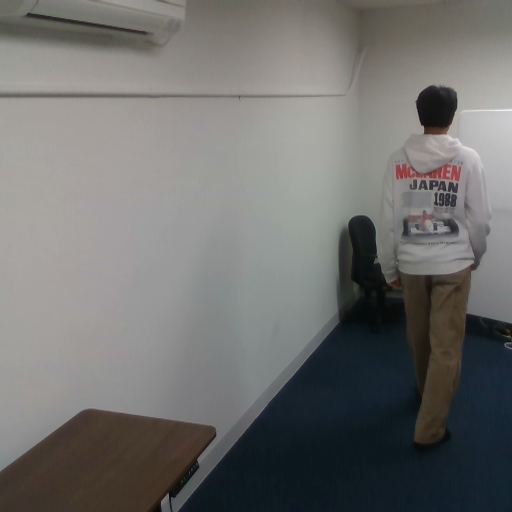} &
    \includegraphics[width=\sdimgscale]{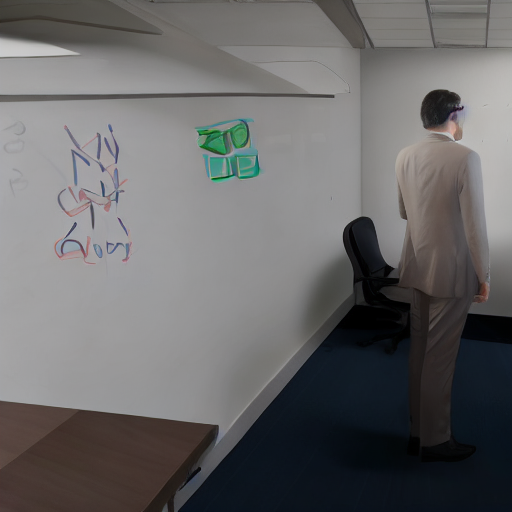} &
    \includegraphics[width=\sdimgscale]{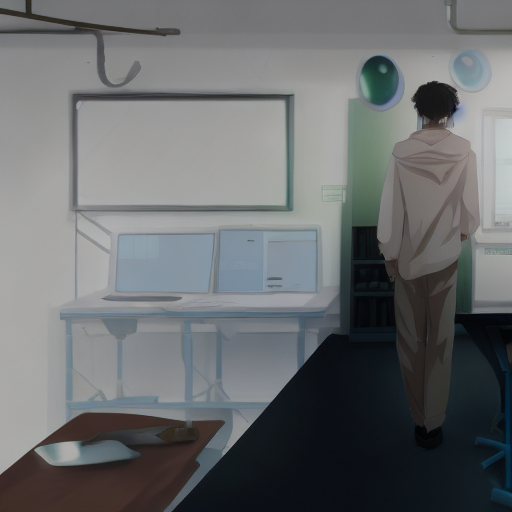} \\

    \raisebox{0.5\dimexpr\sdimgscale}[0pt][0pt]{\rotatebox[origin=c]{90}{Dataset 2}} &
    \includegraphics[width=\sdimgscale]{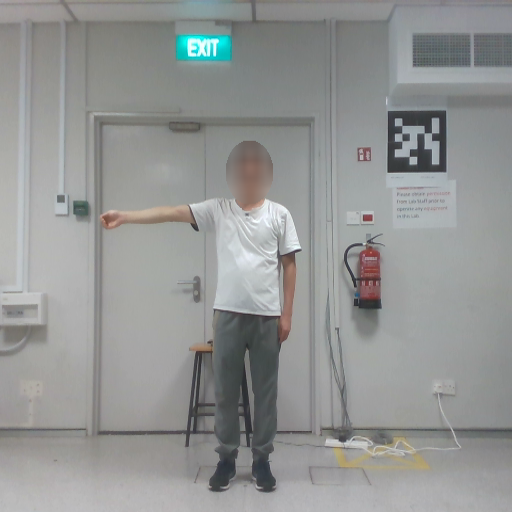} &
    \includegraphics[width=\sdimgscale]{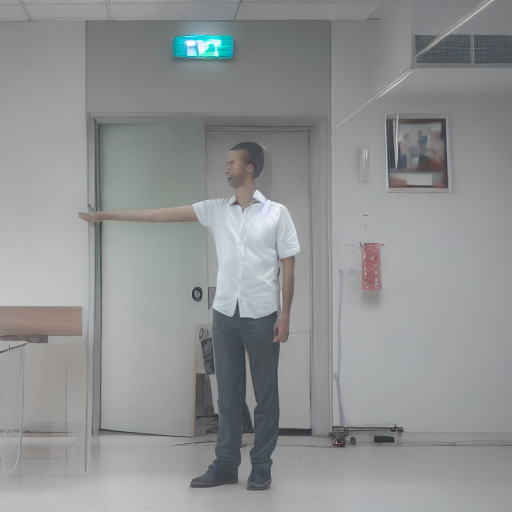} &
    \includegraphics[width=\sdimgscale]{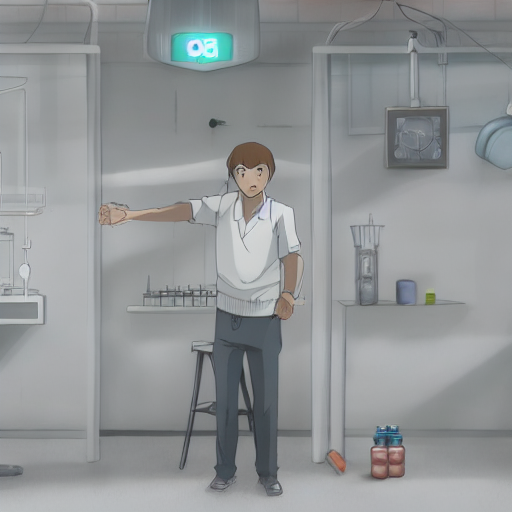} \\
    \bottomrule
  \end{tabular}
\end{table}

\section{Conclusion}
We presented \name, a novel approach for high-resolution image
generation from CSI by leveraging the power of a pretrained
latent diffusion model. Rather than generating images directly in pixel
space—which is computationally expensive and often requires complex
training procedures—we propose a framework that maps CSI amplitudes into the
latent space of the LDM using a lightweight neural network. This design
reduces computational load and simplifies training while
improving perceptual quality compared to conventional end-to-end architectures,
even when the end-to-end model achieves better pixel accuracy. The use
of the LDM's latent space provides a natural process to harness the power of
text-guided image editing using diffusion. This allows for fine-grained control
over the output, which is particularly valuable in privacy-sensitive
contexts. We suggest that \name is a compelling alternative to existing
CSI-to-image methods, offering superior image quality, efficiency, and
controllability.

\renewcommand*{\bibfont}{\normalfont\small}
\printbibliography
\end{document}